\documentclass[a4paper, 10pt, openany]{article}

\usepackage{graphicx}
\usepackage{caption,setspace}
\usepackage{subcaption}
\usepackage{multirow}
\usepackage{float}
\usepackage{longtable}
\usepackage[toc,page,title]{appendix}
\usepackage{gensymb} 
\usepackage{url}
\usepackage[
  separate-uncertainty = true,
  multi-part-units = repeat
]{siunitx}

\usepackage{soul} 
\usepackage{xcolor}
\usepackage{geometry}

\title{\LARGE \bf
Data Augmentation for Leaf Segmentation and Counting Tasks in Rosette Plants
}

\author{Dmitry Kuznichov$^*$, Alon Zvirin$^*$, Yaron Honen$^*$, and Ron Kimmel
\thanks{ Computer Science Department, Technion, Israel Institute of Technology.}%
}

\begin{document}

\maketitle
\thispagestyle{empty}
\pagestyle{empty}


\begin{abstract}

Deep learning  techniques involving image processing and data analysis are constantly evolving. 
Many domains adapt these techniques for object segmentation, instantiation and classification. 
Recently, agricultural industries adopted those techniques in order to bring automation to farmers around the globe. 
One analysis procedure required for automatic visual inspection in this domain is leaf count and segmentation. 
Collecting labeled data from field crops and greenhouses is a complicated task due to the large variety of crops, growth seasons, climate changes, phenotype diversity, and more, especially when specific learning tasks require a large amount of labeled data for training. 
Data augmentation for training deep neural networks is well established, examples include data synthesis, using generative semi-synthetic models, and applying various kinds of transformations.
In this paper we propose a method that preserves the geometric structure of the data objects, thus keeping the physical appearance of the data-set as close as possible to imaged plants in real agricultural scenes. 
The proposed method provides state of the art results when applied to the standard benchmark in the field, namely, the ongoing {\em Leaf Segmentation Challenge} hosted by {\em Computer Vision Problems in Plant Phenotyping}.

\end{abstract}

\section{Introduction}

Visual context, scene understanding, and object location seem to be key factors in image augmentation for deep neural networks. 
There are many ways to augment data in images. 
One of the most prominent ways is cutting objects from the original image, and pasting the objects, exercising geometrical transformations, into a synthetic image. 
Often these operations lead to non-realistic or even non-logical output. 
Gould et al. overcome this problem by understanding the image scene \cite{gould2009decomposing}.
Dvornik et al. find the importance of object locations in the original images  and use these characteristics when deploying the object onto the synthetic image \cite{dvornik2018modeling,dvornik2018importance}.

Several papers dealing with plant phenotyping convey the importance of data augmentation. 
One reason is that training deep neural networks requires a large ground-truth data-set, which is not always available. 
Even if such a data-set exists, augmentation serves to vary the training set, thus improving the learning procedure and performance. 
Moreover, it is a means to train models on synthetic images and enabling the model to generalize and adapt to complex real world scenes \cite{kamilaris2018deep}. 
The importance and contribution of deep learning models for plant phenotyping tasks, especially stress detection and prediction is gaining acceptance and practice. 
Recent surveys on plant phenotyping emphasize the need for data augmentation, and transfer learning in the sense that synthetic data can and should be used for training networks, later tested on real data. 
Main considerations include sufficient amount of balanced data, annotation and normalization of data, and outlier rejection \cite{singh2018deep}. 
Synthetic data modelling, graphical rendering, and transfer learning in context of using pre-trained deep networks (or at least their first layers) for various tasks is also mentioned in other papers dealing with plant genotyping and phenotyping \cite{douarre2018transfer}.

Data augmentation and synthesizing images is gaining acceptance and practice. 
The {\em KITTI} and {\em Cityscapes} datasets are used extensively for semantic understanding of urban scenes \cite{geiger2012we}. 
Some practices are common to all or most approaches (rotations, mirroring, cropping, partitioning, color-space transforms), and can be considered ``general'' while advanced methods are usually applied to specific domains. 
For example, Richardson et al. based their synthesis on parametric modelling of human faces, constructing synthesized models by learning geometric and texture parameters \cite{richardson20163d,richardson2017learning,sela2017unrestricted}. 
Another paper integrates parametric surface modeling with a {\em Generative Adversarial Network} synthesizing realistic facial textures to generate synthetic human faces \cite{slossberg2018high}.
Although applied deep learning is common in analysis of plant structure, and computational and heuristic graphical modeling techniques exist, few attempts have been suggested to combine them. 
Leaf counting and instance segmentation remain a challenge, due to diverse leaf shapes, size and variability during their life cycles in the growth stage, and also due to overlapping and occlusions, abundant number of different crops, and diverse real-world environments (laboratory, greenhouse, field).

Here, we propose a method integrating both approaches, by presenting a method of data augmentation preserving the photo-realistic appearance of plant leaves, and using the augmented data as training set for a network architecture known to achieve high quality results in instance counting and segmentation.
One of the current leading neural networks that developed to solve multiple tasks in the domain of instance segmentation is {\em Mask R-CNN} \cite{he2017mask}. 
This network model is used for object counting, object detection, semantic segmentation and more. 
Designed to tackle multi-category detection and fine segmentation, the network's architecture (evolved gradually) to first propose a large set of probable regions, assign probabilities and continue to segmentation of each object with separate weights for each class. 
The main obstacle to get accurate results is that deep neural networks require a huge annotated dataset which rarely exists. 

We focus on augmenting a plant image training set with photorealistic synthetic images. 
Using a limited amount of images of real leaves and accurate manual segmentation, we use geometric transformations and image processing tools to create a practically infinite amount of synthetic images simulating real-life environments. 
Among these some manipulations can be considered global, like rotations and scaling, while some are tailored specifically for a particular dataset, for example, number of leaves and their orientations in a plant, following a set of formal rules supplemented by random parameter distribution in a reasonable range.

The {\em Computer Vision Problems in Plant Phenotyping} (CVPPP) dataset was created specifically for expected contributions in image based learning related to plant phenotyping \cite{scharr2014annotated,minervini2016finely}.
Originating as the {\em Aberystwyth Leaf Evaluation Dataset} of Arabidopsis Thaliana \cite{bell2016aberystwyth}, the current dataset consists of four (A1 - A4), sets of arabidopsis and tobacco; they contain a separate train/test and also a separate testing subset (A5).
The Rosette image dataset is complemented by two ongoing competitions, introduced in 2014 - the {\em Leaf Segmentation Challenge} and {\em Leaf Counting Challenge} (LSC, LCC respectively), hosted and maintained by CVPPP \cite{PlantPhenotypingDatasets2015}.
Arabidopsis was selected as it is the plant with best known genetics, has a short life span, and a dataset was created in a controlled environment with manual annotations of leaf masks as ground truth. 
Several approaches tackling this dataset are described in \cite{scharr2016leaf}.
Introduced in 2014, the dataset and ongoing challenge already gained considerable impact in plant phenotyping research \cite{tsaftaris2018sharing}.

We also tested our methods on another plant image dataset, collected by Rahan-Meristem \cite{Rahan2018}, as part of a pilot phenotyping project, for future research into early detection of plant stress and prediction of growth stages, in collaboration with the {\em Israel Phenomics Consortium}.
This set contains $50$ images of mature avocado in a plantation and $30$ images of banana plantlets in a greenhouse, with accurate manual segmentation of all leaves. 
Each of these images contain between $20$ to $80$ leaves.

We propose two methods for augmenting an image set by generation of photorealistic synthetic images, preserving geometry and texture as appearing in complex real world agricultural scenes. 
We demonstrate the applicability of these methods to boost a deep neural network performance in accurately counting and segmenting leaves in diverse photographing conditions. 
As an added value, we show some transfer learning ability, in the sense that training on one type of crop images is apparently useful for testing on others. 
Our main contribution is simulation of data to enlarge  the existing data-set with a novel method of synthesizing realistic plant images. 
This process is complemented by using an existing Convolutional Neural Network architecture for the tasks of leaf segmentation and counting. 
Currently, our results are ranked first in both the Leaf Segmentation and Counting Challenges.
In the next section we review several papers concerned with data augmentation aimed specifically for identifying plant parts, especially Rosette leaves. 
The {\em Methods} section describes our approach and strategy for collaging collage  leaf images as means for data augmentation, and the {\em  Results} section presents qualitative and quantitative results on two annotated datasets, the CVPPP Rosette database, and image sets of mature avocado and banana seedling, both with fine manual contouring of leaf masks.

\section{Related Efforts}
 
Taking a deeper inspection at recent efforts focusing on data augmentation by synthesizing leaf images, most draw their ideas from three main approaches: {\em Graphical Modelling}, {\em Domain Randomization} (randomizing parameters such as lighting, pose, textures), and {\em Generative Adversarial Networks}.
Other attempts addressing leaf segmentation and counting rely heavily on neural networks (aimed at image processing tasks), but use a limited augmentation, or train on other datasets, or apply pre-processing, such as color transform, brightness and contrast adjustments,  but not specifically designed for fine contouring of leaf shape nor refined realistic texture.    
Ubbens \& Stavness \cite{ubbens2017deep} introduce an open source platform for plant phenotyping, provide pre-trained networks for common plant phenotyping tasks. 
Recognizing the problem of limited training data that leads to overfitting, they apply brightness and contrast adjustments in addition to cropping and flipping.   

\subsection{Graphical Modelling}
Formalizing plant structure by mathematical models was introduced by Lindenmayer, known as L-systems. 
Formal grammars with a set of rules (functions) are utilized to produce {\em chains} of elements representing plant parts - stems, leaves, roots. 
The process starts with an initial axiom, basic string, applies a set of production rules to create composite strings, and employs a mechanism for transferring the generated string into geometric structures. These models originated in an attempt to assist biological understanding of cell structure and development by formal mathematical models \cite{lindenmayer1968mathematical}. 
Later, these ideas were applied in graphical simulation of plants \cite{prusinkiewicz2012algorithmic}, for rendering synthetic images, and for creating augmented datasets required to train deep neural networks. 
Mundermann et al. empirically model 3D graphical representations of arabidopsis \cite{mundermann2005quantitative}. 
After collecting thousands of measurements of real plants from seedlings to maturity, they infer growth curves of shape, size and position of leaves and stems, and their development over time. 
Although providing insights for key parameters important to formal representation of modeling plant structure and development, their models lack RGB texture, and do not address the issue of plant appearance in images.

Ubbens et al. introduced a parametric version of L-systems for generating synthetic rosettes \cite{ubbens2018use}. 
They simulated growth stages by parametrizing plant components, like apex, leaf, and inter-node, where components are controlled by statistics of leaf length, length/width ratio, leaf age, leaf position and orientation, and number of leaves. 
More important, they demonstrate the capability of ``dataset shift'', training and testing on different datasets, and argue that images of real/synthetic plants are significantly interchangeable when training a neural network. 
Sophisticated leaf positioning is an essential contribution, but their models exhibit small diversity in leaf texture and shape, especially contours. 

\subsection{Domain Randomization}
The main purpose of {\em Domain Randomization} is to tackle the task of object localization, instance detection and possibly object segmentation. 
A few works demonstrate the capability of training entirely on synthesized images, intended for testing on real world scenes. 
This approach intentionally abandons photo-realism by random perturbations of the environment, such as random textures, thus attempting to force the neural network to learn the essential features of the objects \cite{tremblay2018training}.
In practice, this is implemented by developing a simulator producing randomized rendered images. 
The reasoning is that with enough variability in the simulator, images from the real world should appear to the model as just another variation \cite{tobin2017domain}.
{\em Domain Randomization} was originally used for robotic control and automatic navigation, tested on KITTI. 
In the initial setup the randomized parameters were camera and object positions, lighting conditions, and non-realistic textures. 

Applying {\em Domain Randomization} to arabidopsis images is described by Ward et al. \cite{ward2018deep}. 
Their method synthesizes random textures of leaves and background, and constructs separate leaves by deforming a canonical template of a leaf. 
Leaf positions are randomized in a unit sphere, and random camera positions and lighting are applied to produce the images.
The main drawbacks in this approach are that leaves are assumed to be planar, textures have a cartoon-like appearance, and it does not handle background.

\subsection{Generative Adversarial Networks}
Recently achieved popularity for creating realistic image sets, Generative Adversarial Networks (GAN), first introduced by Goodfellow et al. \cite{goodfellow2014generative}, are intended for training neural networks. 
The dual architecture consists of a {\em Generator} and a {\em Discriminator}, trained jointly; the purpose of the {\em Generator} is to produce desired realistic images, while the {\em Discriminator} distinguishes the true from the fake. 
Informally, GANs are known in the deep learning community to be hard to train, but if trained properly generate very good images.
Two recent papers apply {\em conditional Generative Adversarial Networks} (cGAN) to generate artificial images of arabidopsis plants, targeting the Leaf Counting/Segmentation challenge. 
The condition serves as restriction on the training process, is supplemented to the input, and fed to one or more of the networks' layers. 

Giuffrida et al. propose a process starting with random noise, taking number of leaves as the condition concatenated through the networks' layers, ending up with $128\times 128$ pixel RGB images of simulated arabidopsis \cite{giuffrida2017arigan}. 
Deconvolutional layers are part of both the generator and the discriminator components; the discriminator mirrors the generator's architecture, and outputs a binary decision on leaf cardinality. 
The synthetic images resemble an ordered plant-like structure, but leaves appear as floating greenish blobs on a black background, lacking clear contours and veins.

Zhu intruduces a more sophisticatd strategy, first producing masks in a structured manner, and using these masks as input to the GAN \cite{zhu2018data}. 
Individual leaf masks are selected from the ground truth masks, split into $5$ folders by size, and arranged in a logical order by rotations and zooming with small randomization, and placing smaller leaves on top.
These synthesized mask images serve as condition to the generator, which outputs pseudo-real images, replacing leaf masks and mask background with RGB texture.
The discriminator takes an RGB image concatenated to a corresponding mask and distinguishes the real from the fake images.

\subsection{Limited Augmentation}

In addition to the previously mentioned articles, all incorporating data augmentation and synthetic image generation, several other approaches have been applied for segmentation and counting tasks, in particular dealing with the arabidopsis dataset.
Pape \& Klukas used image processing tools for pre-processing - transformation to {\em L*a*b} colorspace and Gaussian blur to assist in separating foreground and background. 
They construct two 3D histograms of {\em L*a*b} values, create a skeleton graph, detect leaf center points, and apply morphological operators and Euclidean distance maps to distinguish between individual leaves \cite{pape20143}. 
In a follow-up paper they used the IAP software for feature extraction, and trained a random forest classifier (utilizing WEKA data minimg software) for leaf border detection \cite{pape2015utilizing}. 

Several attempts using Recurrent Neural Networks (RNN) have been proposed; in short, the basic scheme is iterative, producing a single segmented instance, while updating the current state, stored in a {\em Long Short Term Memory} (LSTM) unit. 
Romera-Paredes \& Torr \cite{romera2016recurrent} suggested an architecture starting  with a CNN to extract image features, and these feature maps are fed to all iterations in the sequence. 
Ren and Zemmel performed dynamic {\em Non Maximal Suppression} to handle occlusions \cite{ren2017end}. Salvador et al. added a encoder-decoder model, the decoder is supplemented with a {\em Convolutional Long Short Term Memory} (ConvLSTM) \cite{salvador2017recurrent}.
All the RNN based attempts apply very limited augmentation, only rotations and flipping of the original images, and report results on the A1 arabidopsis subset only.
A recent RNN based article also reports on collection of a new arabidopsis dataset, time-series image sequences of four accessions under controlled acquisition, in hope and expectation of further research \cite{namin2018deep}.

Other efforts tackle only the counting problem, and treat counting as a direct regression problem, without attempting to segment individual leafs.
Dobrescu et al. use a modified version of Resnet50, applying limited augmentation - rotations, zooming, and flipping of the original images \cite{dobrescu2017leveraging}. 
Aich \& Stavness apply intensity saturation, Gaussian blurr and corresponding sharpening, in addition to rotations and flipping, and use a modified version of SegNet \cite{aich2017leaf}.
Counting leaves by learning features in a non-supervised dictionary learning fashion, without neural networks, was considered by \cite{giuffrida2015learning}. 
Features are extracted from patches of images transformed to a log-polar representation.
The augmentation is a simple shifting of the log-polar representations of the images, analogous to rotations and cropping. 

A note should be made that efforts directed at counting, even if best at guessing number of objects in a snapshot, do not directly address leaf texture nor geometry, therefore having less relevance for plant phenotyping. 
Other attempts, while performing fine positioning of leaves \cite{ubbens2018use,ward2018deep,zhu2018data} generate or rely on exaggerated synthetic leaves, lack real-world textures and diverse geometry, especially leaf contours and their appearance in digital images. 
In the next section, we discuss data augmentation for the specific task at hand, by image synthesis using a model in which leaves are extracted from images of real plant. 

\section{Methods}

Presented here is a method for generation of synthetic images, a technique we term {\em collage}. 
The basic idea is creating a set of segmented leaf images on a transparent background, a single leaf per image, using manual annotations or an automatic procedure. 
In the basic scheme, single leaves undergo geometric transformations with random parameters in a fixed range, and pasted in random locations over selected backgrounds.
The advanced scheme takes into account the logical-semantic relationships among objects, in this case structuring and positioning of leaves as part of the whole plant. 
We apply an algorithm specifically tailored for generating images that seem highly realistic, as if actually taken from the target dataset. 
In case of plants, especially if photoed during a controlled environment, the structuring of leaves as part of the whole plant is important, as well as collaging a photo-realistic image in terms of geometry, texture, occlusions, and background. 

The basic and advanced schemes are termed {\em na\"{i}ve collage} and {\em structured collage}, respectively.
The na\"{i}ve collage is intended for ``in the wild conditions'', when minimal collection of data and annotations are available, and still expecting to do some predictions.
The structured collage exploits certain plant structural attributes assumed or known to be correlated to measurable phenomena. 
We address the following issues, and elaborate on them in the following paragraphs: leaf (object) shape, size, location, ordering and positioning of leaves (object parts) as part of the whole plant, and image background.

\subsection{Na\"{i}ve Collage} 
\begin{figure}[H]
\centering
\begin{subfigure}{.25\textwidth}
  \includegraphics[width=\textwidth]{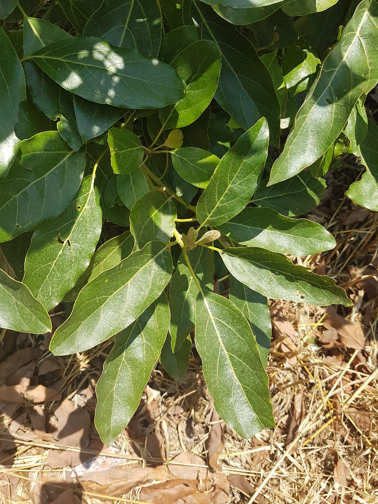}
  \caption{}
  \label{fig:avocado_original}
\end{subfigure}%
$\mkern10mu$
\begin{subfigure}{.25\textwidth}
  \includegraphics[width=\textwidth]{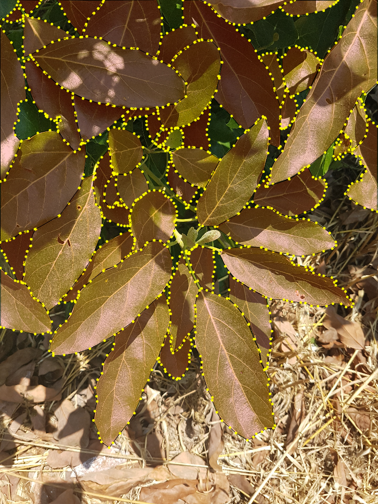}
  \caption{}
  \label{fig:avocado_annotated}
\end{subfigure}%
$\mkern10mu$
\begin{subfigure}{.25\textwidth}
  \includegraphics[width=\textwidth]{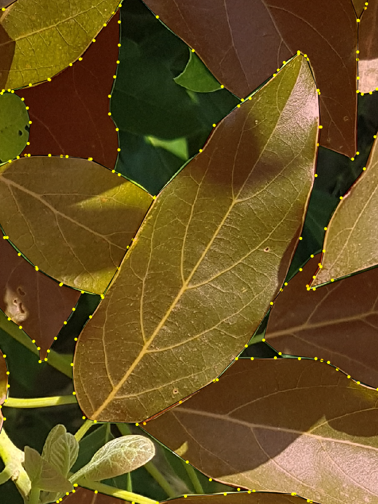}
  \caption{}
  \label{fig:avocado_annotated_zoomed}
\end{subfigure}
\caption{Mature avocado leaves: (a) original image, (b) original with manual annotations, (c) annotation detail. Marked leaves are alpha blended for display purposes.}
\label{fig:originalAvocadoImageExample}
\end{figure}

The na\"{i}ve collage is composed of previously segmented objects on a transparent background. 
The objects are then randomly positioned on selected background images, AS IS, without any logical-semantic relationships among objects. 
At first, this technique was tried on a relatively basic scenario:
The raw data consisted of $\sim$50 high resolution RGB images of avocado leaves ($3000\times 4000$ pixels), supplemented with high quality manual annotations (mask per object) of most  visible leaves (see  Fig. \ref{fig:originalAvocadoImageExample}). 

\begin{figure}[H]
\centering
    \resizebox{.865\textwidth}{!}{$
    \includegraphics[scale=0.1]{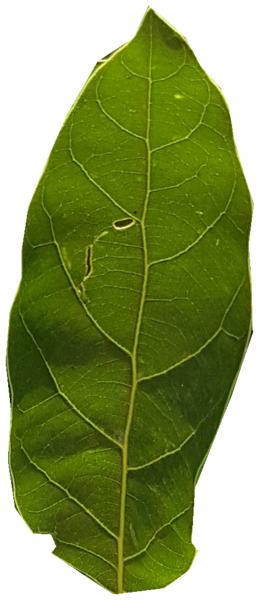}
    \mkern10mu
    \includegraphics[scale=0.1]{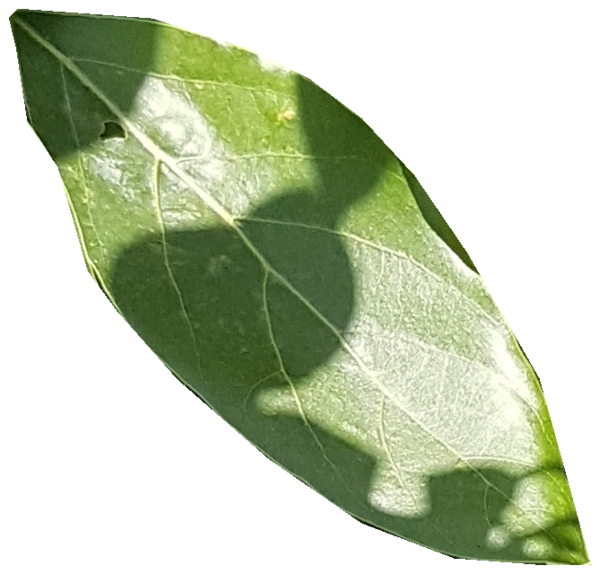}
    \mkern10mu
    \includegraphics[scale=0.1]{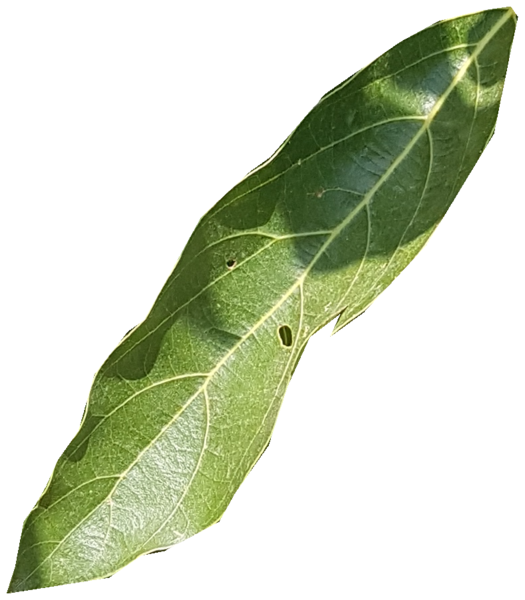}
    \mkern10mu
    \includegraphics[scale=0.1]{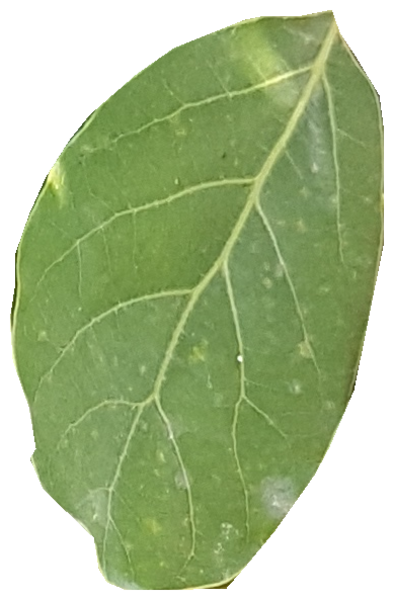}
    $}
\caption{Examples of segmented avocado leaves used for na\"{i}ve collages.}
\label{fig:segmentedAvocadoLeafExamples}
\end{figure}

From the original set of annotated avocado images we extract a set of ``suitable'' leaves, by three criteria: (1) Size (2) Not occluded by other objects (3) Clear and focused appearance. The resulting set consists of $\sim$200 leaves (out of $\sim$3000 original leaves). 
A sample of these finely segmented leaves is displayed in Fig \ref{fig:segmentedAvocadoLeafExamples}. 
The reasoning behind discarding occluded leaves is so the generated image will be as realistic as possible; in the wild it is uncommon to see cut leaves, and the appearance of partial objects is a side effect of the collage algorithm. 
Small and blurry looking leaves are removed due to our deliberate intention of detecting only leaves having a clear and sharp appearance in the image. 
These masked leaves were scaled to $600$ pixels in the largest dimension, preserving the aspect ratio, thus enabling to collage a few dozen leaves in a single $1024\times 1024$ image such that each leaf can be seen clearly. 

\begin{figure}[H]
\centering
\begin{subfigure}{.45\textwidth}
  \includegraphics[width=.9\textwidth]{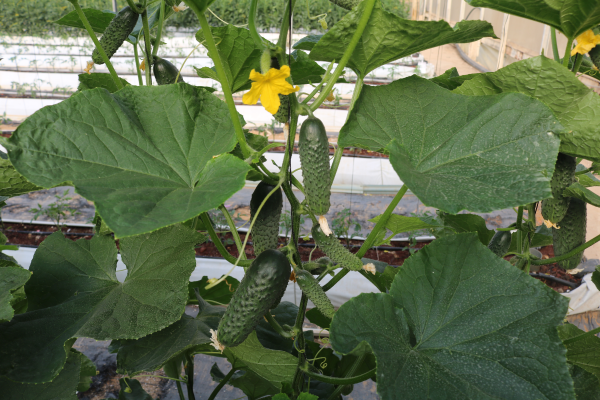}
\end{subfigure}%
\begin{subfigure}{.45\textwidth}
  \includegraphics[width=.9\textwidth]{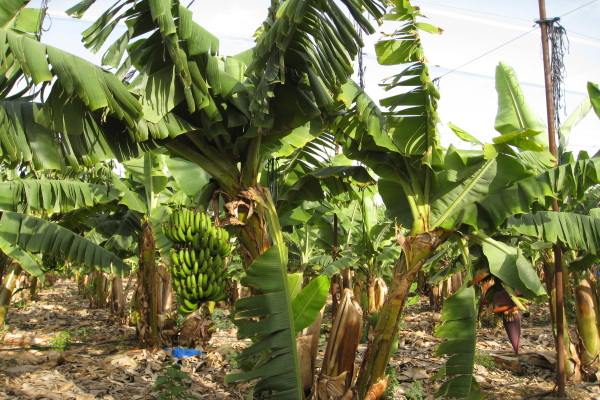}
\end{subfigure}
\caption{Examples of agricultural scenes used as background for na\"{i}ve collages.}
\label{fig:naiveCollageBackgroundImageExample}
\end{figure}

As background for the synthesized images we used cropped images of size $1024\times 1024$ from a set of $24$ high resolution agricultural images, not including clearly seen avocado leaves (see Figure \ref{fig:naiveCollageBackgroundImageExample}). 
The rationale for background selection is based on the intention of accurately detecting fine looking objects, ( i.e., leaves of a certain crop), and distinguish them from other botanical objects (stems, fruit, ground), mostly appearing in images as a composition of green shades.

\begin{figure}[H]
\centering
\begin{subfigure}{.85\textwidth}
    \centering
    \includegraphics[height=4.5cm]{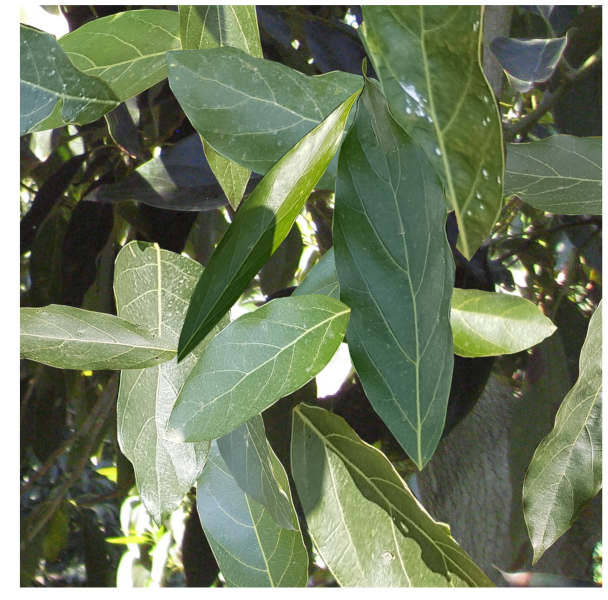}
    \includegraphics[height=4.5cm]{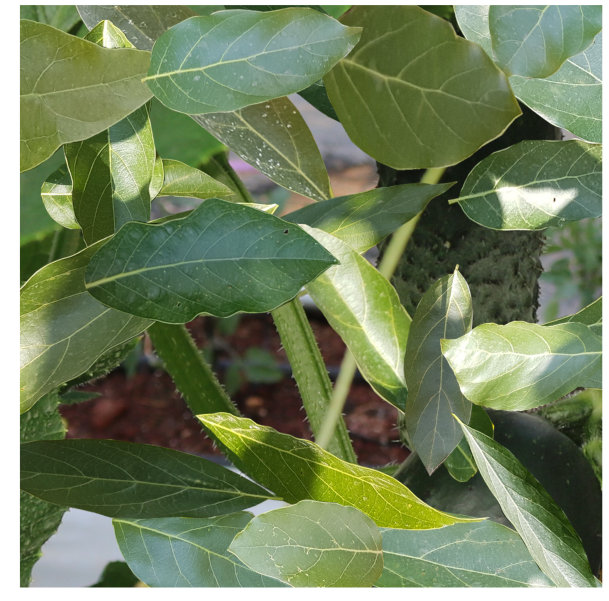}
    \caption{}
    \label{fig:naive_collage_images}
\end{subfigure}
\begin{subfigure}{.85\textwidth}
    \centering
    \includegraphics[height=4.5cm]{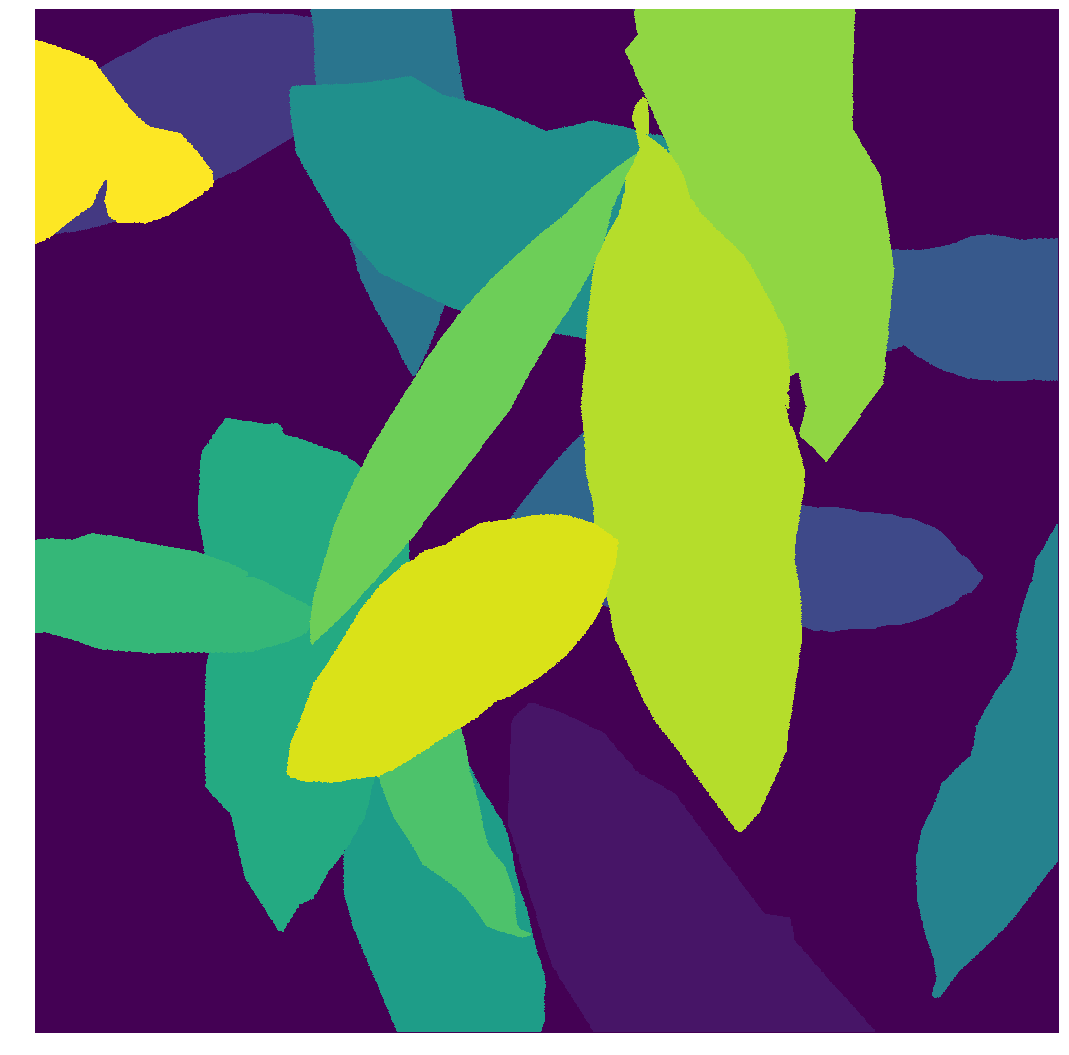}
    \includegraphics[height=4.5cm]{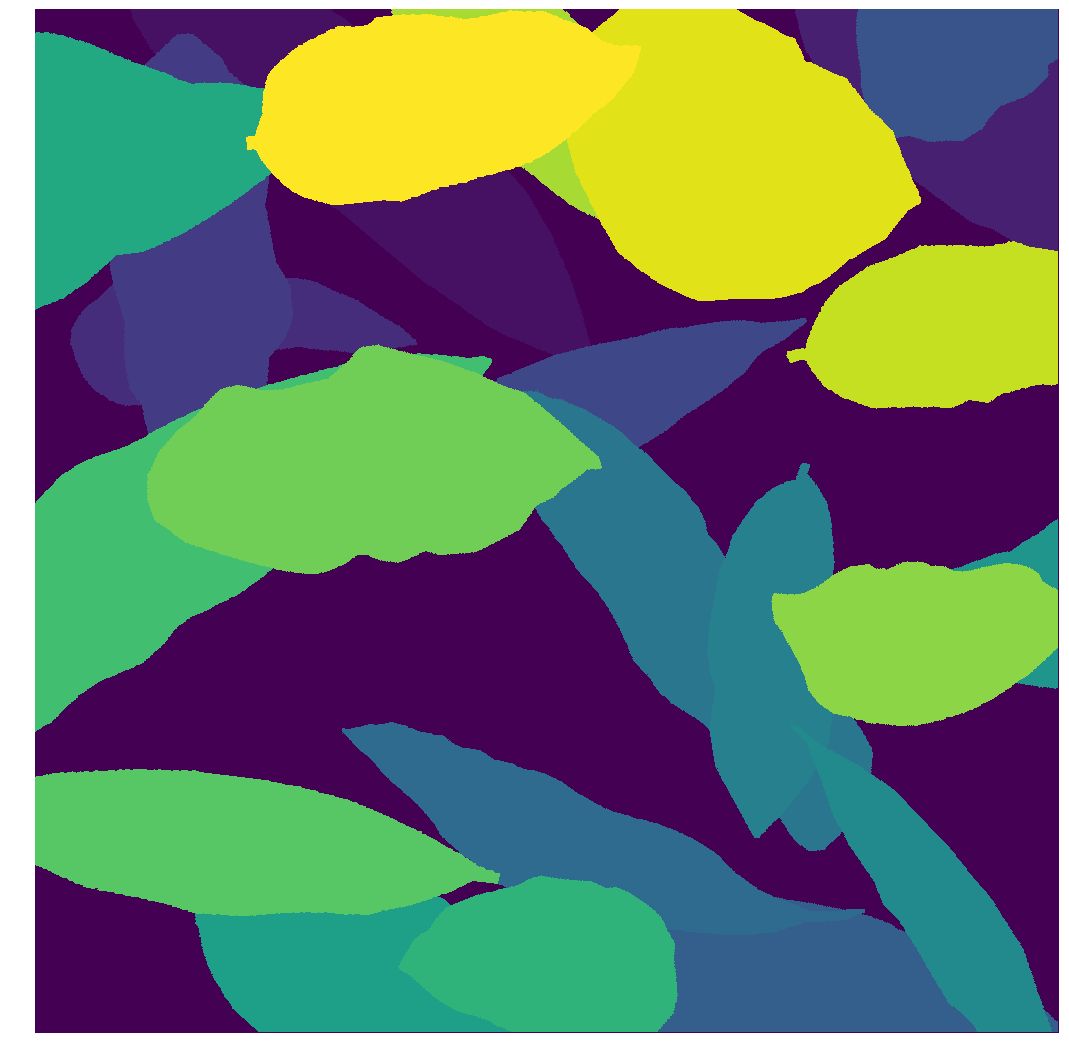}
    \caption{}
    \label{fig:naive_collage_masks}
\end{subfigure}
\caption{Generated images (a) and corresponding masks (b) from the avocado training set.}
\label{fig:augmentedImageNaiveCollageExample}
\end{figure}

The collage is created by positioning $10$ to $40$ segmented leaves (random number in arbitrary but fixed range) in random locations/scale and rotations on top of a background image. 
The location of each leaf is randomly selected, the only restriction is that the leaf center remains inside the background image ($1024\times 1024$). 
Horizontal and vertical scaling are independent, with random values between $0.4$ and $1.1$. 
The rotation angle of each leaf (with respect to its original orientation) is randomly selected in the range $0-359$ degrees. 
We did not apply affine or projective transformations, in order to preserve the original point of view, although it can be done as well. 
Parallel to image creation, we generate corresponding masks fitting the created image; examples are displayed in Figure \ref{fig:augmentedImageNaiveCollageExample}.

\subsection{Structured Collage}
\begin{figure}[H]
    \centering
    \includegraphics[width=.85\textwidth]{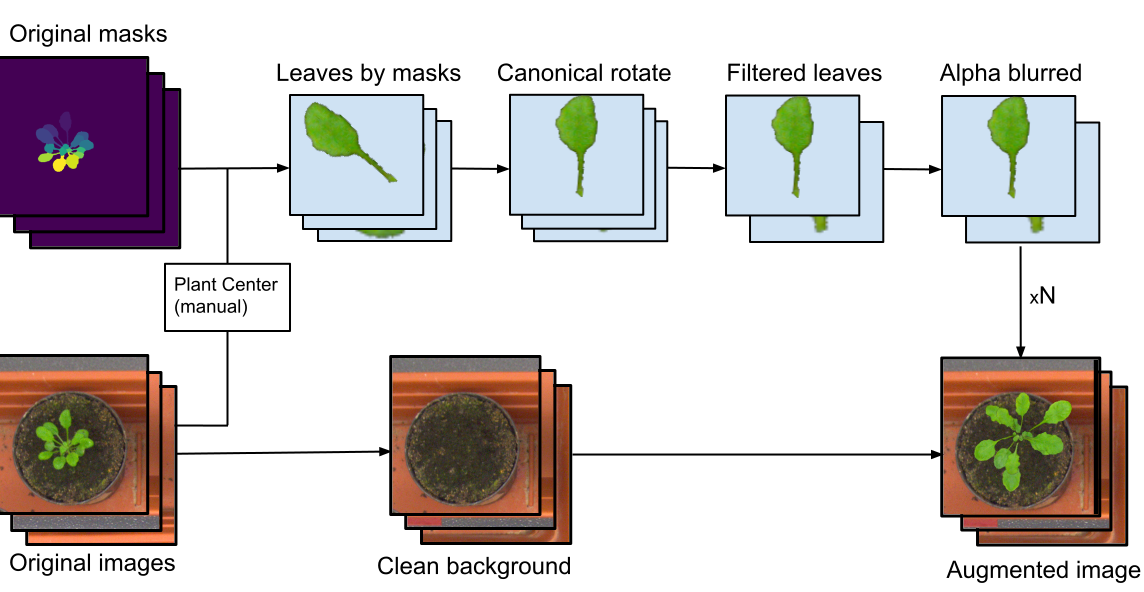}
    \caption{Structured collage generation pipeline.}
    \label{fig:structured_collage_pipeline}
\end{figure}

The extended collage version takes into account the logical order, structure, and hierarchical relationship among objects placed in the image, specifically the location, size and shape of individual leaves as part of the whole plant. 
This is especially important in case of specific datasets such as the arabidopsis and tobacco images, all photoed from above, capturing plant structure at progressive development stages. 
It should be noted that although we present a description for synthesizing images with appearance akin to this specific dataset, similar steps, with fine tuning of parameters, can be employed for datasets based on different plant species or other image acquisition systems. 
In short, the process for collage generation consists of selecting an appropriate background, creating a set of alligned leaves in cannonical form, and logical insertion of leaves from this set onto the image. 
The workflow is depicted in Figure \ref{fig:structured_collage_pipeline}; details and considerations of this collaging process are described in the following paragraphs.

\subsubsection{Background images}

\begin{figure}[H]
    \centering
    \resizebox{.85\textwidth}{!}{$
    \includegraphics[height=3cm]{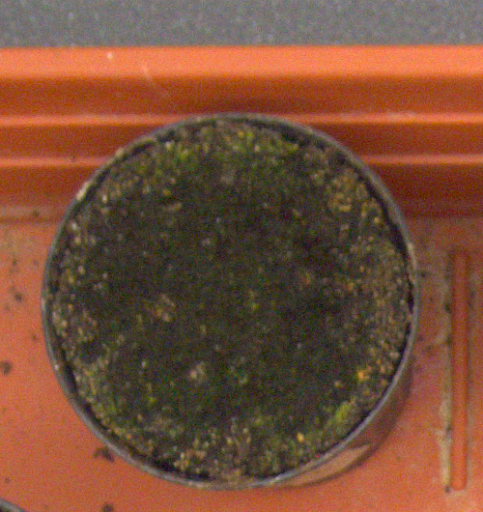}
    \mkern1mu
    \includegraphics[height=3cm]{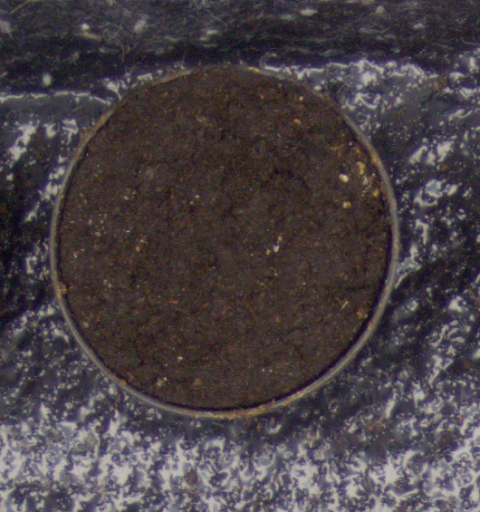}
    \mkern1mu
    \includegraphics[height=3cm]{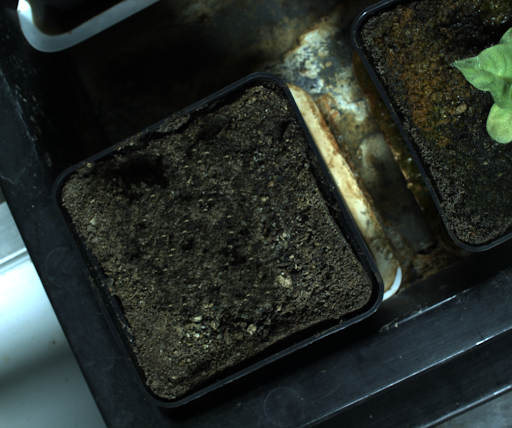}
    \mkern1mu
    \includegraphics[height=3cm]{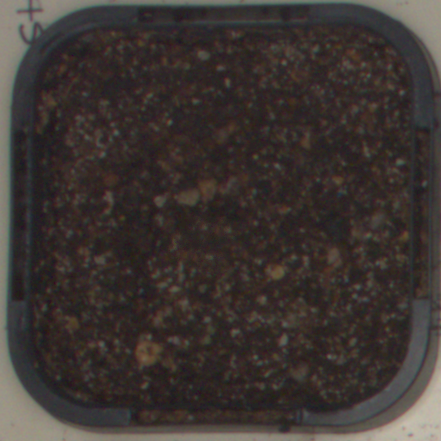}
    $}
    \caption{Clean background images, extracted from the A1-A4 Rosette subsets.}
    \label{fig:backgroundImageExampleSmart}
\end{figure}

As a first step we created $112$ background images, using original images from the dataset and applying a semi-manual segmentation of the plant from the background by activation of a heal-selection filter \cite{GIMP2018}. 
In total, $16, 26, 26,$ and $44$ background images were created, matching the A1, A2, A3 and A4 subsets, respectively. 
Examples are displayed in Figure \ref{fig:backgroundImageExampleSmart}.

\subsubsection{Leaves in canonical form}

\begin{figure}[H]
\centering
\begin{subfigure}[b]{.2\textwidth}
    \centering
    \includegraphics[height=2cm]{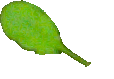}
    \caption{}
    \label{fig:aligned_leaf_a}
\end{subfigure}
\begin{subfigure}[b]{.2\textwidth}
    \centering
    \includegraphics[height=2cm]{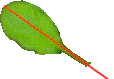}
    \caption{}
    \label{fig:aligned_leaf_b}
\end{subfigure}
\begin{subfigure}[b]{.2\textwidth}
    \centering
    \includegraphics[height=3.65cm]{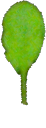}
    \caption{}
    \label{fig:aligned_leaf_c}
\end{subfigure}
\caption{Arabidopsis leaf: (a) original, (b) with principal axis originating in plant center, (c) aligned in canonical form.}
\label{fig:leafStepsSmartCollage}
\end{figure}

In the next step all leaves were cut from the training set images according to their masks, and rotated to align them in canonical form. 
The rotation angle was determined as the angle between the horizontal axis and the leaf's principal axis, defined as a segment connecting the plant center (as manually marked) and the farthest mask pixel from the center. 
In all the aligned images the central pixel in the bottom row corresponds to the plant center. 
An example of original leaf image, principal axis and aligned form are depicted in Figure \ref{fig:leafStepsSmartCollage}.

As a result $11096$ aligned leaf images were created: $2088, 287, 146,$ and $8575$ in the A1, A2, A3 and A4 subsets, respectively. 
However, many of these leaves are not suitable for generation of realistic images; the main criteria for retaining leaves are: (1) the leaf mask contains no more than one component, (2) the leaf base is not too far from the plant center, and (3) the leaf appears fully (or almost) in the image, with minimal occlusion.

\begin{figure}[H]
    \centering
    \resizebox{.45\textwidth}{!}{$
    \includegraphics[height=1cm]{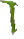}
    \mkern20mu
    \includegraphics[height=2cm]{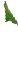}
    \mkern20mu
    \includegraphics[height=2cm]{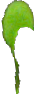}
    \mkern20mu
    \includegraphics[height=2cm]{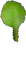}
    \mkern20mu
    \includegraphics[height=2cm]{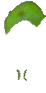}
    $}
    \caption{Examples of arabidopsis leaves discarded from the generation procedure.}
    \label{fig:badLeavesExampleSmart}
\end{figure}

Examples of discarded leaves are shown in Fig \ref{fig:badLeavesExampleSmart}.
After removal, $5883$ ($\sim50\%$) were left: $1363, 219, 102$ and $4199$ in the A1, A2, A3 and A4 categories, respectively. 
Note that although these leaves are discarded from the aligned set used in the generation procedure, similar looking partial and occluded leaves are expected to be detected and masked. 
The generation procedure, described in the next section, produces just this kind of occlusions, as well as full leaves.

\subsubsection{Synthetic image generation}

Since our goal is to create realistic images, we heuristically attempt to imitate the plant's structure, using visual observations as rules of thumb:
(1) The plant spawns from a point near the pot's center. 
(2) All the leaves grow from this point towards the periphery. 
(3) Leaf size and distance from the center are correlated. 
(4) Angles between leaves follow a certain distribution.

The main idea of image generation is the same for all datasets although each of the four subsets (A1-A4) should have its set of parameters fine-tuned. 
These include number of leaves, plant center, and inter-leaves angle randomization. 
The synthesized images are generated according to the following steps:
\begin{enumerate}
  \item Randomly select a background image from the background image list of the specified set.
  \item Randomly select number of leaves for the image (value in a fixed range).
  \item Randomly select the plant center coordinates, up to maximal value from the image center.
  \item Add first leaf (from the aligned set) with a random rotation angle.
  \item For each additional leaf: 
  \begin{enumerate}
    \item select leaf angle based on previously added leaves. (last added leaf on top.)
  \end{enumerate}
\end{enumerate}

In a formalized manner, let us define:
\\ \(l_j \) as leaf \( j \) in a \( n \)-leaf dataset \(L=\{l_j\}_{j=1}^n\), 
\\ \(I_k \) as background image \(k\) from the background set,
\\ \(I_k^i \) as background image \(k\) with i added leaves, (note \(I_k  = I_k^0\), by definition),
\\ and \(T_i (I,l) \) as an operator adding leaf \(l\) to image \(I\) as \(i^{th}\) leaf in the image. (note \(i \geq 1\)).

The synthesized image is initialized as \(I_k^0\). 
\\Adding leaf \(j\) to background image \(k\) containing \(i-1\) previously added leaves results in image \(I_k^i\), specified by
\begin{eqnarray*}
I_k^i \leftarrow T_i(I_k^{i-1}, l_j).
\end{eqnarray*}
Since the process is iterative, it follows that,
\begin{eqnarray*}
 I_k^i &\leftarrow& T_i(T_{i-1}(I_k^{i-2}, l_{j'}), l_j)
\cr
I_k^i &\leftarrow& T_i(\cdots T_1(I_k^0, l_{j_1}) \cdots , l_j).
\end{eqnarray*}
The operator \(T_i\) is dependant on a few parameters, namely leaf location and rotation.
\(T_1\) is initialized with a plant center location \((x,y)\)  and a first leaf angle \(\alpha_1 \); values of these parameters are randomly chosen from a fixed range and remain constant till the end of a single image creation,
\begin{eqnarray*}
T_1 \Leftarrow ((x,y), \alpha_1).
\end{eqnarray*}
Obviously, while trying to simulate plant structure by object collaging, \(T_i\)'s angle \(\alpha_i\) is a function of number of objects and all angles of previously added objects, and can be defined iteratively by
\begin{eqnarray*}
\alpha_i&=&f(\alpha_{i-1}, i).
\end{eqnarray*}
Taking a look at the Rosette dataset \cite{PlantPhenotypingDatasets2015}, we notice approximately \(120\degree\) between consecutive leaves, and similar to \cite{ubbens2018use}, a basic formulation of \(\alpha_i\) can be stated as
\begin{eqnarray*}
\alpha_i &=& \alpha_{i-1} + \SI{127.5\pm 12.5}{\degree}
\end{eqnarray*}
Observing that Rosette leaves grow in triads, with slight modification, first triad as before \(\alpha_i = \alpha_{i-1} + \SI{127.5\pm 12.5}{\degree} \), and first leaf (only) of each new triad \SI{60\pm 10}{\degree}, \SI{30\pm 5}{\degree}, etc.

\begin{figure}[H]
    \centering
    \resizebox{.65\textwidth}{!}{$
    \includegraphics[height=3cm]{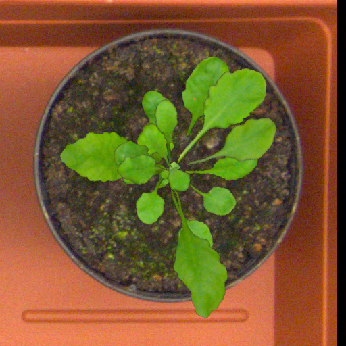}
    \mkern5mu
    \includegraphics[height=3cm]{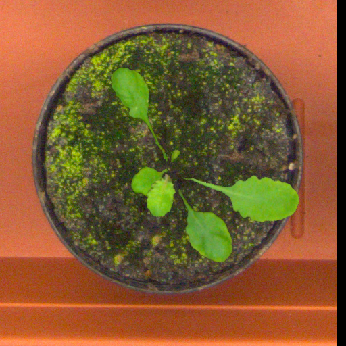}
    \mkern5mu
    \includegraphics[height=3cm]{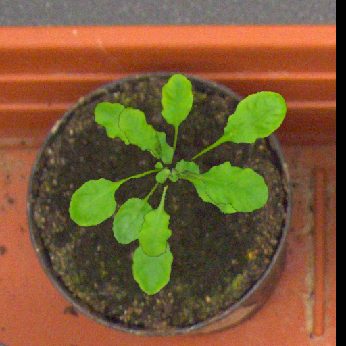}    
    $}
    \\[0.5mm]
    \resizebox{.65\textwidth}{!}{$
    \includegraphics[height=3cm]{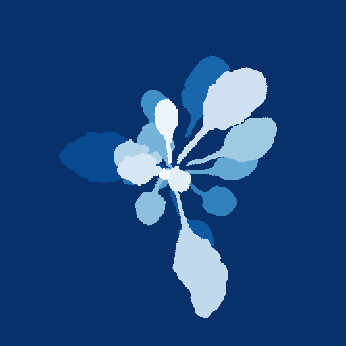}
    \mkern5mu
    \includegraphics[height=3cm]{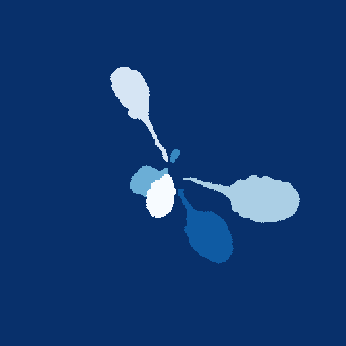}
    \mkern5mu
    \includegraphics[height=3cm]{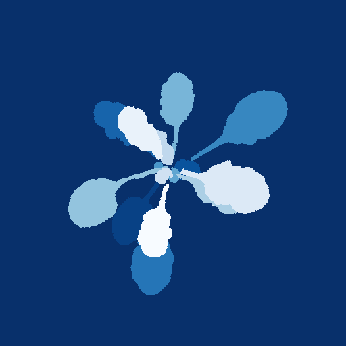}
    $}
    \caption{Examples of generated images and masks simulating the arabidopsis A1 subset.}
    \label{fig:Synthetic_Images_A1_Example}
\end{figure}

Parameters that were chosen for each dataset can be found in Table \ref{table:datasetParameters}.
The size of the training images is restricted to multiples of $64$ due to memory alignment. 
Plant center locations are  from a threshold range surrounding the image center.  
The number of leaves provides reasonable range for leaf cardinality in each subset.
Examples of synthesized  images of arabidopsis plants simulating the A1 subset are displayed in Figure \ref{fig:Synthetic_Images_A1_Example}.

\begin{table*}[!h]
\centering
\caption{}\centering
\begin{tabular}{|l|l|l|l|l|l|}
\hline
\begin{tabular}[c]{@{}l@{}}Data\\ set\end{tabular} & 
\begin{tabular}[c]{@{}l@{}}Original\\ image size\end{tabular} &
\begin{tabular}[c]{@{}l@{}}Train\\ image size\end{tabular} &
\begin{tabular}[c]{@{}l@{}}Image\\ center\end{tabular} &
\begin{tabular}[c]{@{}l@{}}Plant center \\ delta\end{tabular} &
\begin{tabular}[c]{@{}l@{}}Number\\ of leaves\end{tabular} \\ \hline
 A1 & 530x500 & 512x512 & (256,256) & 40x40 & 5-25\\ \hline
 A2 & 530x565 & 512x512 & (256,256) & 40x40 & 3-25\\ \hline
 A3 & 2448x2048 & 2048x2048 & (1024,1024) & 160x160 & 2-15\\ \hline
 A4 & 441x441 & 448x448 & (224,224) & 35x35 & 4-30\\ \hline
 \end{tabular}
\label{table:datasetParameters}
\end{table*}

\section{Results}

Testing was performed on the CVPPP dataset (all categories, including A5), as well as the avocado and banana sets. 
The na\"{i}ve collage is used in the avocado/banana case, more suitable as ``in the wild" conditions, where images were taken at various camera positions and include a variety of light conditions. 
The structured collage is aimed at the CVPPP dataset, acquired under controlled, consistent conditions, and exhibiting coherent plant structure.
The network was trained on this type of generated images and masks. 
We perform both the counting and segmentation tasks with the publicly available Matterport implementation \cite{Matterport2018} of Mask-R-CNN, pre-trained on the COCO dataset.

We kept the default hyper-parameters, except the following: BATCH\_SIZE  = 2, IMAGES\_PER\_GPU = 2, and STEPS\_PER\_EPOCH = 10.  
The first two parameters were modified due to GPU memory limitation, and the last one to save the network's weights (after each epoch Mask R-CNN saves all the weights) to examine the learning progress step by step. 

\subsection{Na\"{i}ve Collage}
\begin{center}
\begin{figure}[H]
\centering
\resizebox{.65\textwidth}{!}{$
    \includegraphics[scale=0.1]{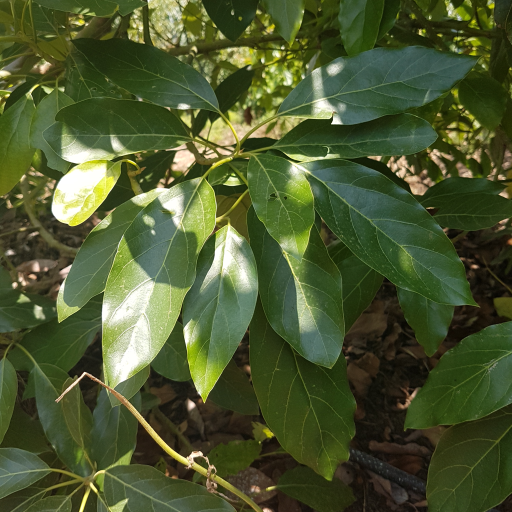}
    \mkern5mu
    \includegraphics[scale=0.1]{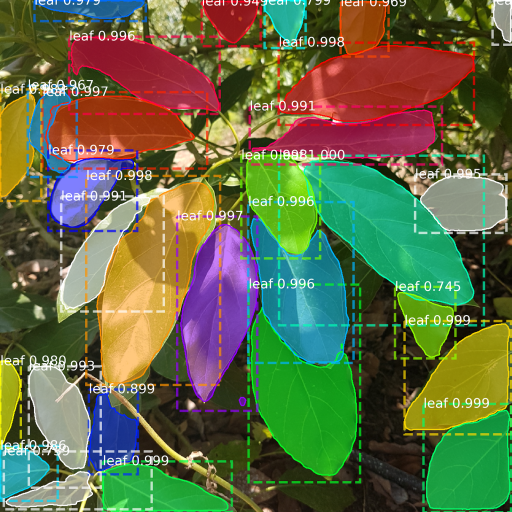}
$}
\caption{Avocado test image: original and Mask R-CNN segmentation result.}
\label{fig:AvocadoLeafDetectionDemo}
\end{figure}
\end{center}

\begin{figure}[H]
    \centering
    \resizebox{.85\textwidth}{!}{$
    \includegraphics[height=3cm]{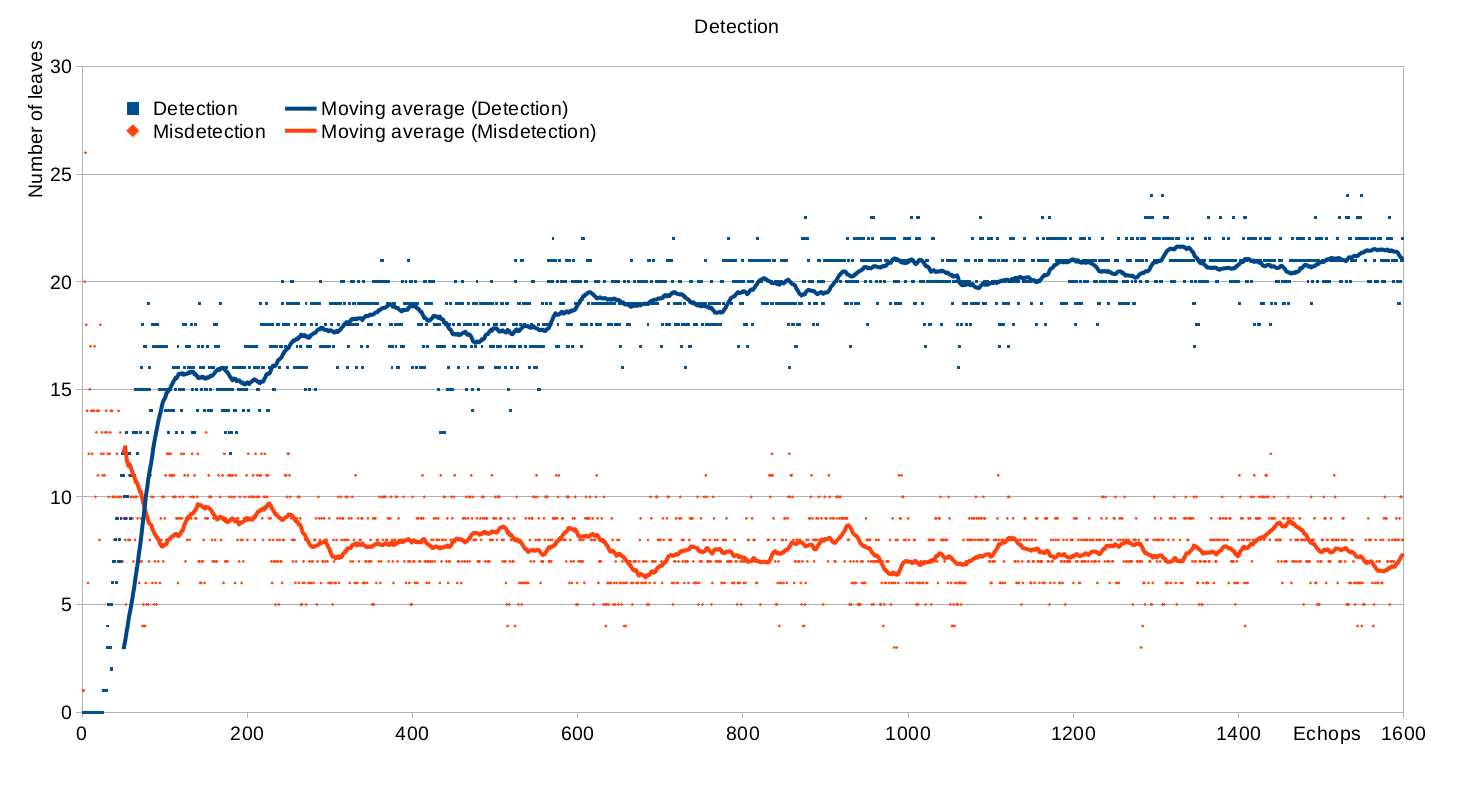}
    $}
\caption{Detection vs misdetection rates on avocado leaves, by training epochs.}
\label{fig:AvocadoLeafDetectionGraph}
\end{figure}

To test the networks performance we use a single image from the manually annotated dataset, leaves of which were separated from the training set. 
These leaves are expected to be segmented by the network, after training on collaged images. 
We decided to evaluate correct segmentation by comparing leaf area for all leaves in the test image having mask areas over a fixed threshold (700 pixels). 
For correct segmentation we assume at least 0.8 IOU (Intersection over Union) of leaf area with respect to its manually annotated counterpart - mask area of the ground truth. 
Mask R-CNN shows a clear pattern of instance segmentation improvement on the test image during progression of training epochs. 
Instance segmentation is visualized in Figures \ref{fig:AvocadoLeafDetectionDemo} and \ref{fig:AvocadoLeafDetectionGraph} shows number of leaves rightly/wrongly segmented, according to training epochs. 

\begin{figure}[H]
    \centering
    \resizebox{.85\textwidth}{!}{$
    \includegraphics[height=3cm]{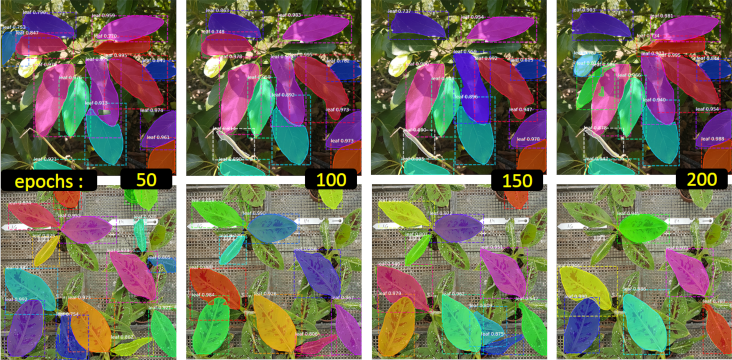}
    $}
\caption{Avocado and banana leaf detection: segmentation results by training epochs. Training is performed on synthesized images of avocado only.}
\label{fig:BananasaFromAvocado}
\end{figure}

A few other possibilities were examined, based on results obtained with the na\"{i}ve collage. 
We tested the network's capability of transfer learning - training on collages of one crop and achieving good results on another. 
Figure \ref{fig:BananasaFromAvocado} visualizes segmentation results on images of avocado and banana leaves, after training on avocado only. 
It seems that after $50$ epochs the network achieves high accuracy of avocado leaf detection. 
Banana leaf detection at this point is reasonable as well; as the learning process continues detection of bananas leaves starts to deteriorate. 
A possible explanation is that in the first epochs region proposals for leaves in general are probably good, resulting in fine mask detection of both avocado and banana leaves.
After that the network gained specialized knowledge specific to avocado leaves, therefore rejecting banana leaves as a legitimate category.

\subsection{Structured Collage}
As detailed in the {\em Methods} part, the network was trained on structured collage images. 
Although the collages are generated from extracted leaves and corresponding masks, we did not use the original images, nor transformations of the originals, in the training process.
For validation we used the original training data of the four subsets A1 - A4, containing ground truth instance segmentations per leaf. 
We did not use the leaf centers, nor the foreground/background masks of the whole plant, also supplied as part of the ground truth training data. 

Submissions to the Leaf Segmentation and Counting Challenges (LSC and LCC) are published in the CodaLab website \cite{CodaLab2019}, and are evaluated by four criteria,
\begin{itemize}
\item BestDice - Dice coefficient between leaf segmentation results and the ground truth.
\item AbsDiffFG - Absolute difference between number of leaves in the results and the ground truth. 
\item DiffFG - Difference between number of leaves in the results and the ground truth. 
\item FgBgDice - Dice coefficient between foreground/background mask segmentation results and the ground truth. 
\end{itemize}

\begin{figure}[H]
\centering
\begin{subfigure}{.25\textwidth}
  \includegraphics[width=\textwidth]{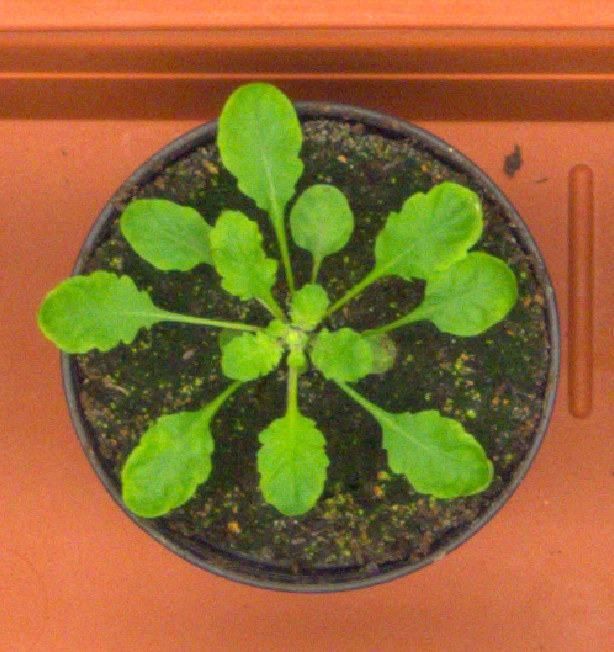}
  \caption{}
  \label{fig:SmartCollageMaskResult_a}
\end{subfigure}%
$\mkern10mu$
\begin{subfigure}{.25\textwidth}
  \includegraphics[width=\textwidth]{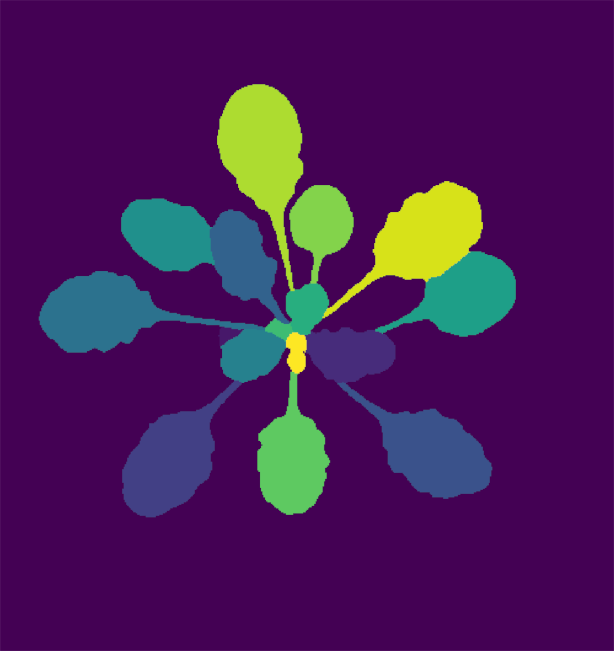}
  \caption{}
  \label{fig:SmartCollageMaskResult_b}
\end{subfigure}%
$\mkern10mu$
\begin{subfigure}{.25\textwidth}
  \includegraphics[width=\textwidth]{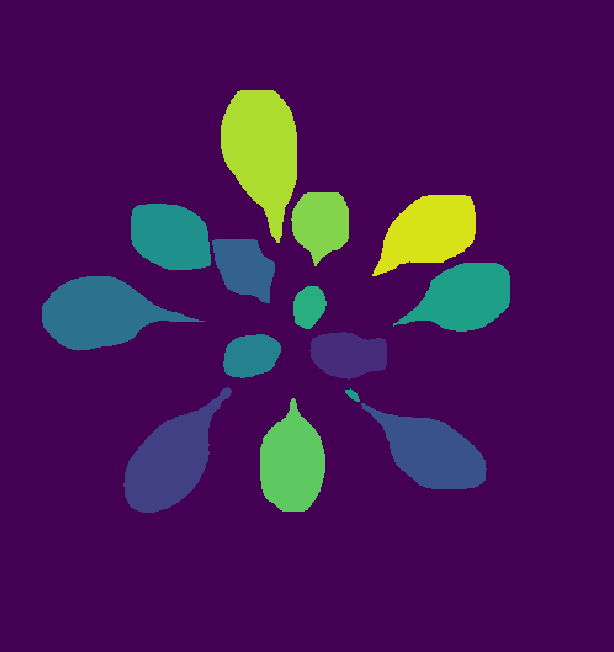}
  \caption{}
  \label{fig:SmartCollageMaskResult_c}
\end{subfigure}
\caption{Structured Collage result example: (a) training image, (b) training mask, (c) network segmentation of the image.}
\label{fig:SmartCollageMaskResult}
\end{figure}

A full description of the evaluation can be found in \cite{cvppp2017data_description}. 
Since our main goal is accurate leaf contouring, decision was to focus on BestDice results; the Dice coefficient is equal by definition to IOU (Intersection over Union) evaluation. 
First we validate the network's performance with the evaluation script supplied by the challenge, training each subset separately.
This allows us to choose the best epoch for each subset and run the test with this epoch's weights. 
Visualization of a training example from the arabidopsis A1 category, its ground-truth leaf masks and the network's segmentation results are presented in Figure \ref{fig:SmartCollageMaskResult}. 

\begin{figure}[H]
    \centering
    \resizebox{.85\textwidth}{!}{$
    \includegraphics[height=3cm]{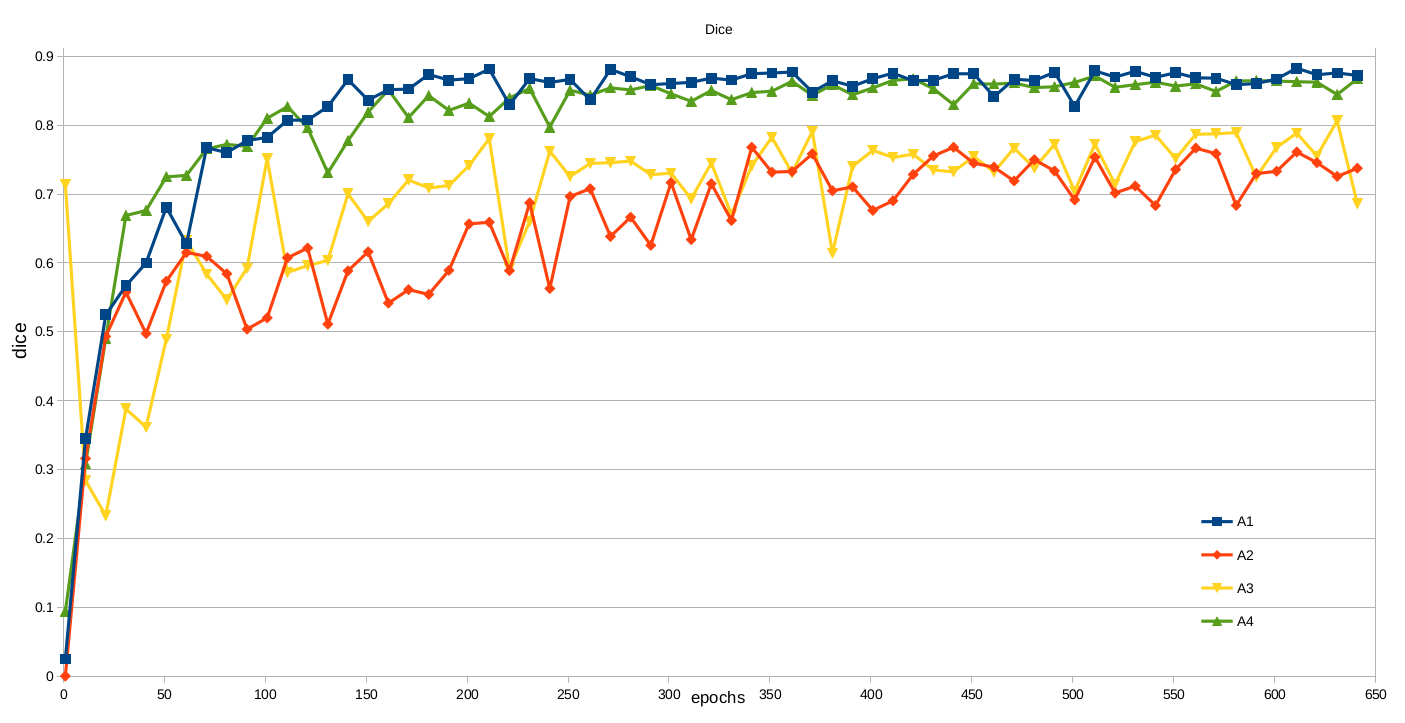}
    $}
    \caption{BestDice evaluation on training data on the 4 categories of the dataset.}
    \label{fig:validation Dice}
\end{figure}

\begin{figure}[H]
    \centering
    \resizebox{.85\textwidth}{!}{$
    \includegraphics[height=3cm]{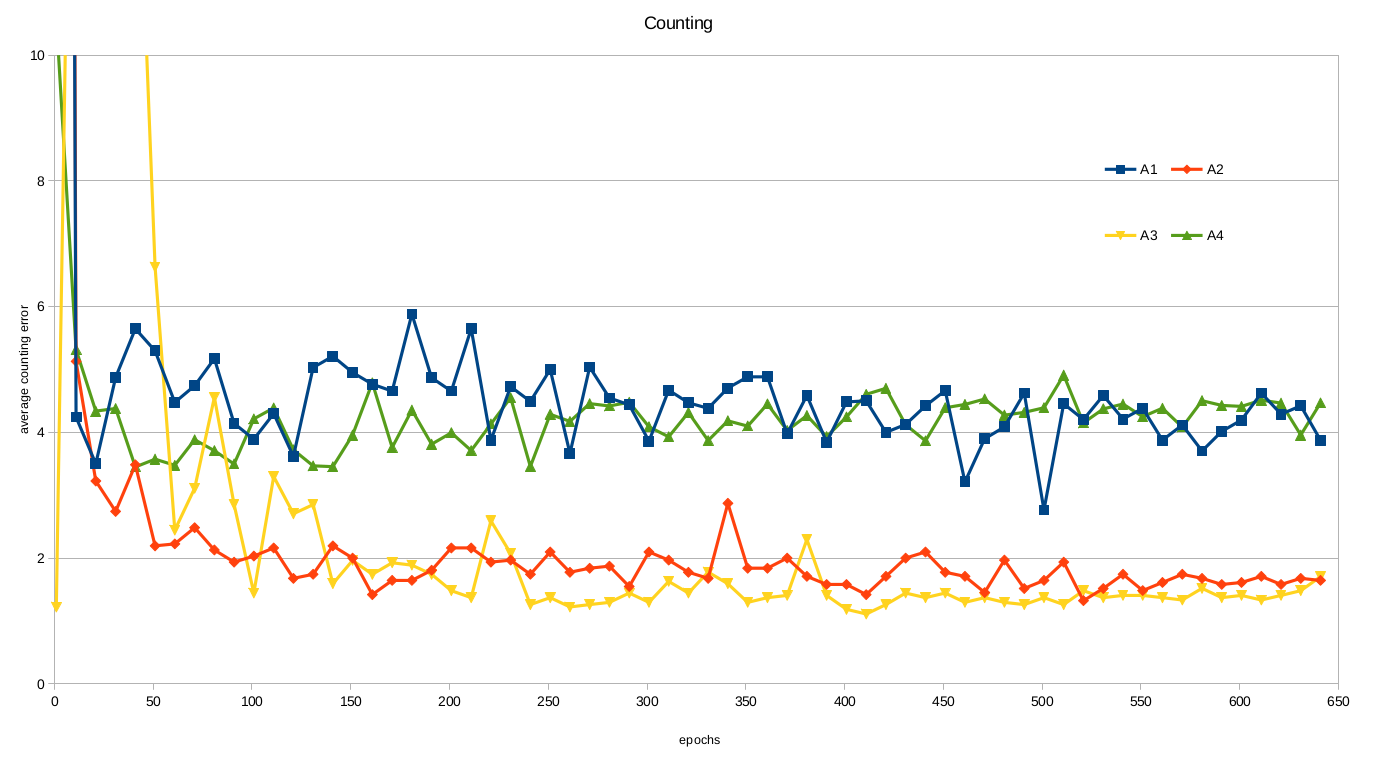}
    $}
    \caption{Absolute difference of counting error on the 4 categories of the dataset.}
    \label{fig:validationAbsDiffFG}
\end{figure}

The network's performance, by training epochs, is presented in Figure \ref{fig:validation Dice} (BestDice evaluation) and Figure \ref{fig:validationAbsDiffFG} (absolute difference in count). 
Although we expect to see some over-fitting in the evaluation graph (recalling that training and validation data are different), the network's performance remains stable from epoch 400, as can be seen in both figures. 
Evidently, scores on the A2 and A3 subsets are poorer in segmentation accuracy and better at counting, compared to A1 and A4.
A possible explanation is the smaller sets of extracted leaves from the A2,A3 categories that the network was trained on.
Since the optimization loss is a combination of five functions, this scenario leads the network learning, while struggling to improve segmentation loss, to direct its efforts on count loss.

For testing, we chose the epoch (weights) scoring best on the  validation set. 
As for test results (calculated by CodaLab \cite{CodaLab2019}), we are currently ranked first in the A2,A3,A4,A5 and mean over all subsets for Dice, and second in the A1 subset. 
In the counting challenge we are ranked first in the A2,A3,A4,A5 and mean over all subsets. 
Our submitted results are presented in Table \ref{table:CodaLab_Evaluation_Scores}.

\begin{table*}[!h]
\centering
\caption{CodaLab evaluation scores}
\centering
\begin{tabular}{|l|l|l|l|l|l|l|}
\hline
 & A1 & A2 & A3 & A4 & A5 & Mean  \\ \hline
 BestDice & 0.878 & 0.835 & 0.801 & 0.873 & 0.851 & 0.854\\ \hline
 AbsDiffFG & 5.06 & 3.44 & 1.64 & 4.48 & 3.979 & 3.95\\ \hline
 DiffFG & -5.06 & -3.44 & -1.5 & -4.48 & -3.936 & -3.91\\ \hline
 FgBgDice & 0.886 & 0.837 & 0.899 & 0.873 & 0.837 & 0.859\\ \hline
 \end{tabular}
\label{table:CodaLab_Evaluation_Scores}
\end{table*}

\section{Conclusion}

We have shown that data augmentation preserving geometric features and sophisticated positioning of objects enhances network performance in the tasks of object instance detection and segmentation. 
The suggested method was tested on the publicly available dataset of Rosette plants, and achieved highest scores on the leaf segmentation task, and is currently ranked first in leaf counting as well. 
As an added benefit we demonstrate that training by structured generation of synthesized images not only achieves good results on a specific plant crop, but is transferable to accurate segmentation of other crops, and of images acquired in different conditions. 
We hope this modest contribution will serve to motivate further investigation of integrating synthetic data augmentation with real world botanical scenes for various plant phenotyping tasks.




\section*{Acknowledgment}
This research was partly supported by the Israel Innovation Authority, the Phenomics Consortium.

{\small
\bibliographystyle{unsrt}
\bibliography{arabidopsis}
}

\end{document}